\documentclass{llncs}
\usepackage{amsmath} 
\usepackage{amssymb} 

\usepackage{graphicx}
\usepackage{subfigure}
\usepackage{siunitx}

\begin{document}

\title{X-ray-transform Invariant Anatomical Landmark Detection for Pelvic Trauma Surgery}

%
%
\author{Bastian~Bier\inst{1,2,}\thanks{These authors have contributed equally and are listed in alphabetical order.}, Mathias~Unberath\inst{2,*}, Jan-Nico~Zaech\inst{1,2}, Javad~Fotouhi\inst{2}, Mehran~Armand\inst{3}, Greg~Osgood\inst{4}, Nassir~Navab\inst{2,*} \and Andreas~Maier~\inst{1,*}}
\authorrunning{Bier and Unberath et al.} 
%
%
\institute{Pattern Recognition Lab, Friedrich-Alexander-Universit{\"a}t Erlangen-N{\"u}rnberg
\and
Computer Aided Medical Procedures, Johns Hopkins University
\and 
Applied Physics Laboratory, Johns Hopkins University
\and
Department of Orthopaedic Surgery, Johns Hopkins Hospital
}

\maketitle              

\begin{abstract}
X-ray image guidance enables percutaneous alternatives to complex procedures. Unfortunately, the indirect view onto the anatomy in addition to projective simplification substantially increase the task-load for the surgeon. Additional 3D information such as knowledge of anatomical landmarks can benefit surgical decision making in complicated scenarios. Automatic detection of these landmarks in transmission imaging is challenging since image-domain features characteristic to a certain landmark change substantially depending on the viewing direction. Consequently and to the best of our knowledge, the above problem has not yet been addressed.\\
In this work, we present a method to automatically detect anatomical landmarks in X-ray images independent of the viewing direction. To this end, a sequential prediction framework based on convolutional layers is trained on synthetically generated data of the pelvic anatomy to predict 23 landmarks in single X-ray images. View independence is contingent on training conditions and, here, is achieved on a spherical segment covering \ang{120}$\times$\ang{90} in LAO/RAO and CRAN/CAUD, respectively, centered around AP. On synthetic data, the proposed approach achieves a mean prediction error of $5.6\pm 4.5$\,mm. We demonstrate that the proposed network is immediately applicable to clinically acquired data of the pelvis. In particular, we show that our intra-operative landmark detection together with pre-operative CT enables X-ray pose estimation which, ultimately, benefits initialization of image-based 2D/3D registration. 

\keywords{Landmark detection, Orthopedics, ConvNets, X-ray}
\end{abstract}

\section{Introduction} 

X-ray image guidance during surgery has enabled percutaneous alternatives to complicated procedures reducing the risk and discomfort for the patient. This benefit for the patient comes at the cost of an increased task-load for the surgeon, who has no direct view on the anatomy but relies on indirect feedback through X-ray images. These suffer from the effects of projective simplification; particularly the absence of depth cues, and vanishing anatomical landmarks depending on the viewing direction. Therefore, many X-rays from different views are required to ensure correct tool trajectories~\cite{Dose,CArmMoving}. Providing additional, ''implicit 3D'' information during these interventions can drastically ease the mental mapping from 2D images to 3D anatomy~\cite{KWireMental,UseOfNavigation}. In this case, implicit 3D information refers to data that is not 3D as such but provides meaningful contextual information related to prior knowledge of the surgeon.\\
A promising candidate for implicit 3D information are the positions of anatomical landmarks in X-ray images. Landmark or key point detection is well understood in computer vision, where robust feature descriptors disambiguate correspondences between images, finally enabling purely image-based pose retrieval. Unfortunately, the above concept defined for reflection imaging does not translate directly to transmission imaging, since the appearance of the same anatomical landmark can vary substantially depending on the viewing direction. Consequently and to the best of our knowledge, X-ray-transform invariant anatomical landmark detection has not yet been investigated. However, successful translation of the above concept to X-ray imaging bears great potential to aid fluoroscopic guidance.\\
In this work, we propose an automatic, purely image-based method to detect anatomical landmarks in X-ray images independent of the viewing direction. Landmarks are detected using a sequential prediction framework~\cite{CPM} trained on synthetically generated images. Based on landmark knowledge, we can a) identify corresponding regions between arbitrary views of the same anatomy and b) estimate pose relative to a pre-procedurally acquired volume without the need for calibration. We evaluate our approach on synthetic validation data and demonstrate that it generalizes to unseen clinical X-rays of the pelvis without the need of re-training. Further, we argue that the accuracy of our detections in clinical X-rays may benefit the initialization of 2D/3D registration.\\
While automatic approaches to detect anatomical landmarks are not unknown in literature, all previous work either focuses on 3D image volumes~\cite{Ghesu} or 2D X-ray images acquired from \emph{a single predefined} pose~\cite{CephalometricChallenge,Chen14}. In contrast to the proposed approach that restricts itself to implicit 3D information, several approaches exist that introduce explicit 3D information. These solutions rely on external markers to track the tools or the patient in 3D~\cite{markelj2012review}, consistency conditions to estimate relative pose between X-ray images~\cite{aichert2015epipolar}, or 2D/3D registration of pre-operative CT to intra-operative X-ray to render multiple views simultaneously~\cite{markelj2012review,tuckertowards}. While these approaches have proven helpful, they are not widely accepted in clinical practice. The primary reasons are disruptions to the surgical workflow~\cite{UseOfNavigation}, susceptibility to both truncation and initialization due to the low capture range of the optimization target~\cite{hou2017predicting}.

\section{Method}

\subsection{Background}

\begin{figure} [tb]
  \centering
  \includegraphics[width=1.0\textwidth]{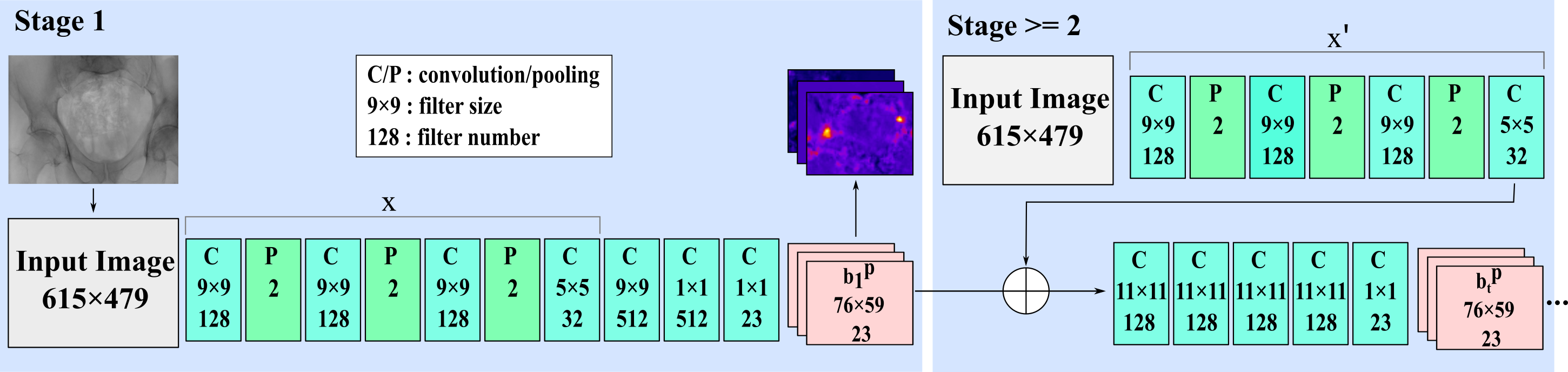}
  \caption{Network architecture: In subsequent stages, landmarks are predicted from belief maps of the previous stage and image features of the input image. Adapted from~\cite{CPM}.}
  \label{fig:network}	
\end{figure}

Recently, sequential prediction frameworks proved effective in estimating human pose from RGB images~\cite{CPM}. The architecture of such network is shown in Fig.~\ref{fig:network}. Given a single image, the network predicts belief maps $b_t^p$ for each anatomical landmark $p\in \{1,...,P\}$. The core idea of the network is that belief maps are predicted in stages $t\in \{1,...,T \}$ using both local image information and long-range contextual dependencies of landmark distributions given the belief of the previous stage. 
In the first stage of the network, the initial belief $b_1^p$ is predicted only based on local image evidence $x$ using a block of convolutional and pooling layers. In subsequent stages $t\geqslant2$, belief maps are obtained using local image features $x'$ and the belief maps of the preceding stage. Over all stages, the weights of $x'$ are shared. The cost function is defined as the sum of L2-distances between the ground truth $b_*^p(z)$ and the predicted belief maps accumulated over all stages. The ground truth belief maps are normal distributions centered around the ground truth location of that landmark. The network design results in the following properties: (1) In each stage, the predicted belief maps of the previous stage can resolve ambiguities that appear due to locality of image features. The network can learn that certain landmarks appear in characteristic configurations only. (2) To further leverage this effect, each output pixel exhibits a large receptive field on the input image of $160\times160$. This enables learning of implicit spatial dependencies between landmarks over long distances. (3) Accumulating the loss over the predicted belief in multiple stages diminishes the effect of vanishing gradients that complicates learning in large networks.


\subsection{X-ray Transform Invariant Landmark Detection}

We exploit the aforementioned advantages of sequential prediction frameworks for the detection of anatomical landmarks in X-ray images independent of their viewing direction. Our assumption is that anatomical landmarks exhibit strong constraints and thus characteristic patterns even in presence of arbitrary viewing angles. In fact, this assumption may be even stronger compared to human pose estimation if limited anatomy, such as the pelvis, is considered due to rigidity. Within this paper and as a first proof-of-concept, we study anatomical landmarks on the pelvis. We devise a network adapted from~\cite{CPM} with six stages to simultaneously predict 23 belief maps per X-ray image that are used for landmark location extraction (Fig.~\ref{fig:network}). Implementation was done in tensorflow, with a learning rate of 0.00001, and a batchsize of one. Optimization was performed using Adam over 30 epochs (convergence reached).

\paragraph{Landmark Detection} Predicted belief maps $b_t^p$ are averaged over all stages prior to estimating the position of the landmarks yielding the averaged belief map $b^p$. We define the landmark position $l_p$ as the position with the highest response in $b^p$. Landmarks with responses $b^p<0.4$ are discarded since they may be outside the field of view or not reliably recognized. 

\paragraph{Training Data}

Training data is synthetically generated from full body CTs from the NIH Cancer Imaging Archive~\cite{dataset}. In total, 20 CTs (male and female patients) were cropped to an ROI around the pelvis and 23 anatomical landmark positions were annotated manually in 3D. Landmarks were selected to be clinically meaningful and clearly identifiable in 3D; see Fig.~\ref{fig:beliefMaps} a). From these volumes and 3D points, projection images and projected 2D positions were created, respectively. X-rays had $615\times479$ pixels with an isotropic pixel size of $0.616$\,mm. The belief maps were downsampled eight times. During projection generation augmentation was applied: We used random translations in all three axes, variation of the source-to-isocenter distance, and horizontal flipping of the projections. Further and most importantly, we sampled images on a spherical segment with a range of \ang{120} in LAO/RAO and \ang{90} in CRAN/CAUD centered around AP, which approximates the range of variation in X-ray images during surgical procedures on the pelvis~\cite{khurana2014pelvic}. The forward projector computes material-dependent line integral images, which are then converted to synthetic X-rays. A total of 20.000 X-rays with corresponding ground truth belief maps were generated. Data was split $18\times1\times1$-fold in training, testing, and validation. We ensured that images of one patient are not shared among sets.

\section{Experiments and Results}

\subsection{Synthetic X-rays}

\begin{figure}[tb] 
\centering
{\includegraphics[width=1.0\textwidth]{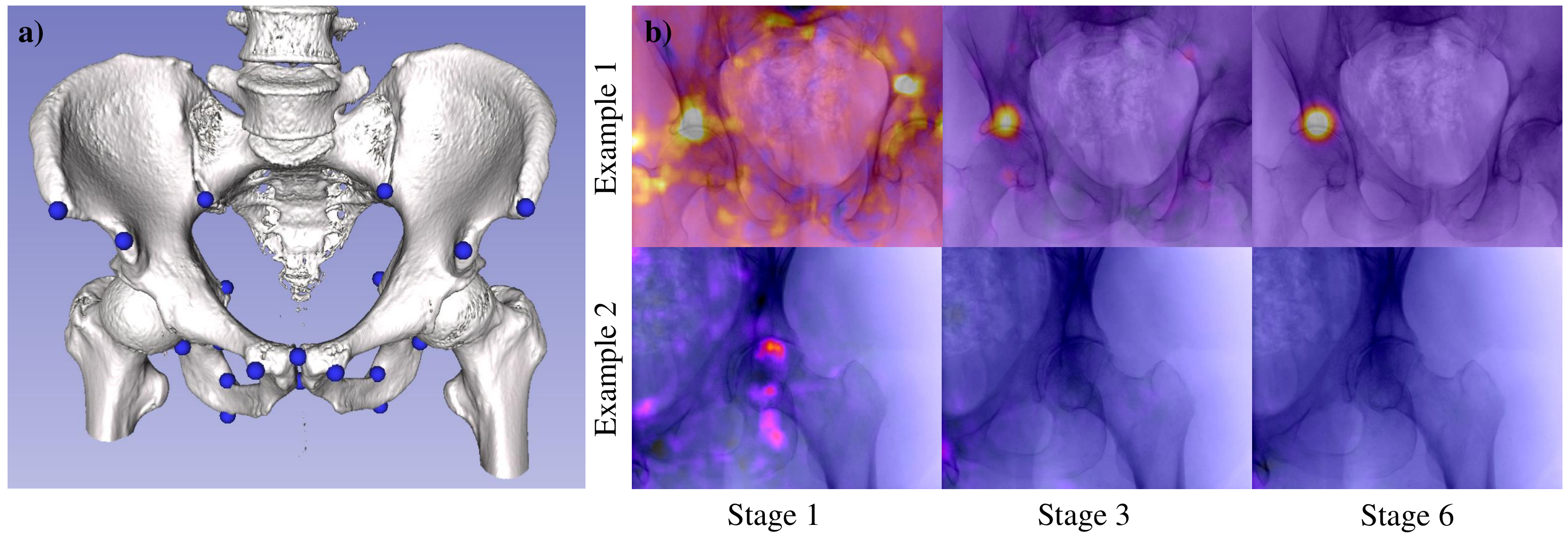}} 
\caption{Uncertainty in the local feature detection after stage 1 is resolved in subsequent stages. This is shown for a symmetric response (Example 1, anterior inferior iliac spine) and for a landmark not in the FOV (Example 2, tip of femoral head).} 
\label{fig:beliefMaps}
\end{figure}

\begin{figure}[tb] 
\centering
\includegraphics[width=1.0\textwidth]{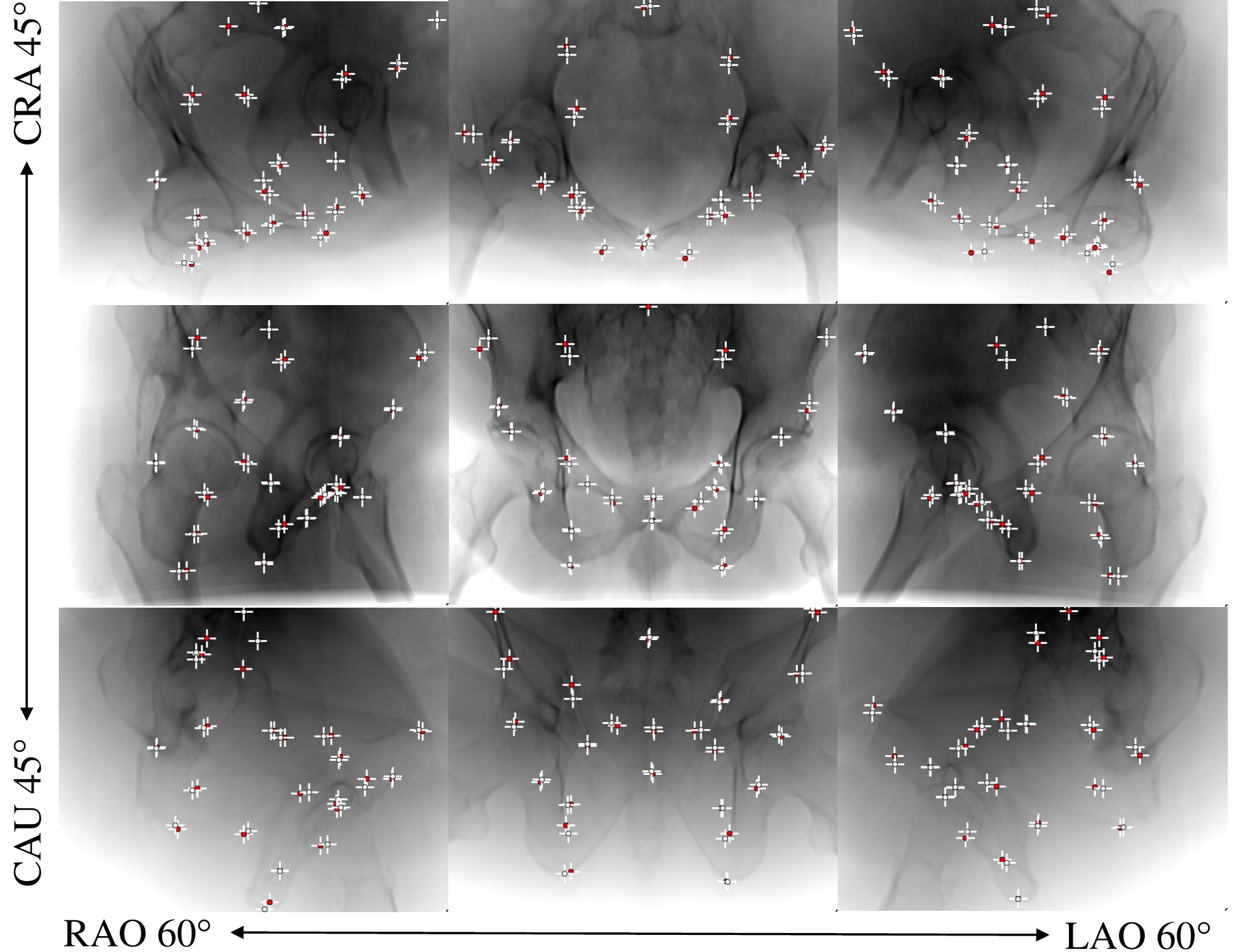}
  \caption{Detection results over the sampled sphere. White and red marker positions indicate ground truth and predicted landmark location, respectively.}
  \label{fig:wallOfIndependence}	
\end{figure}

\begin{figure}[tb] 
\centering
\subfigure{\includegraphics[height=0.3\textwidth]{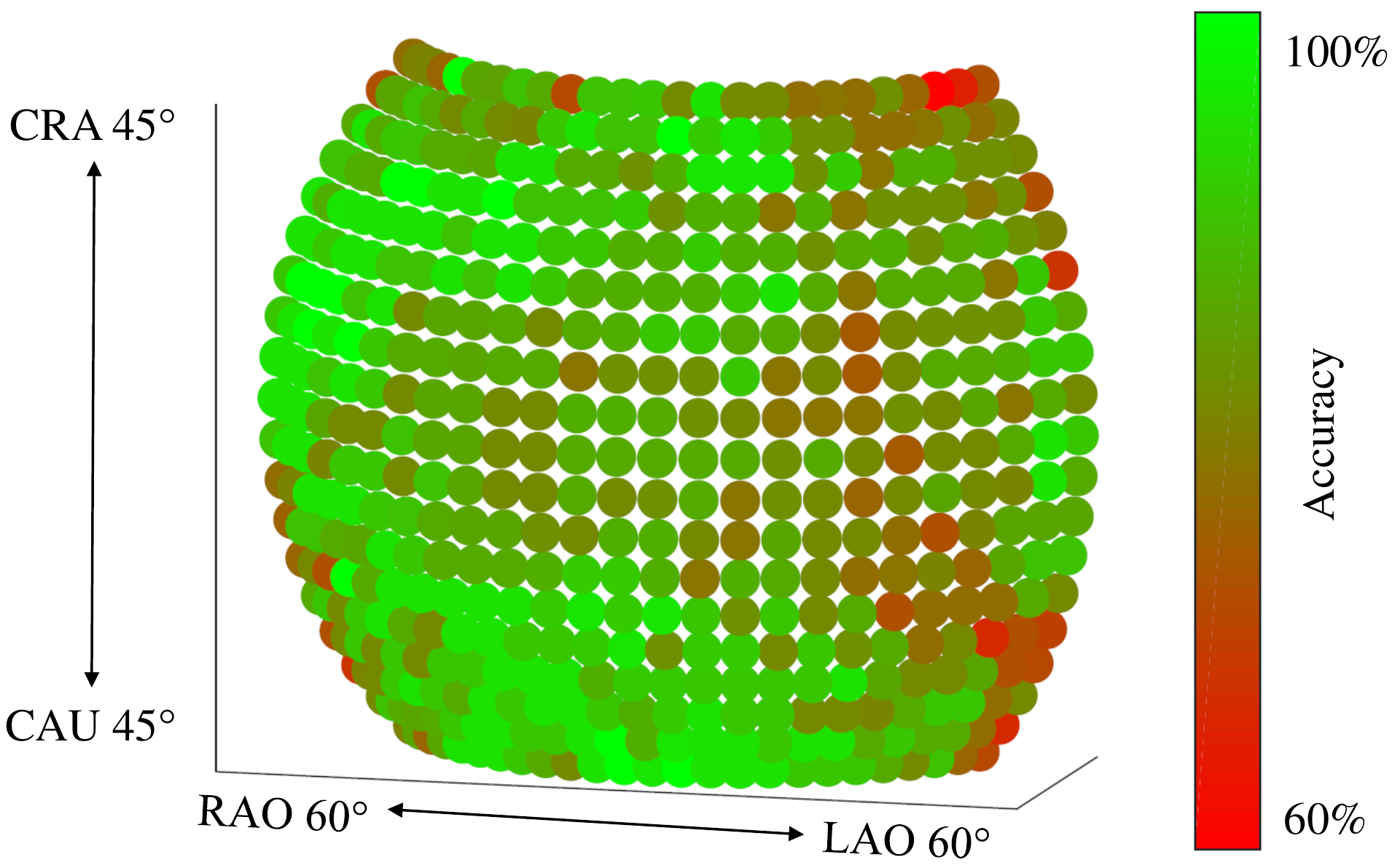}} 
\subfigure{\includegraphics[height=0.3\textwidth]{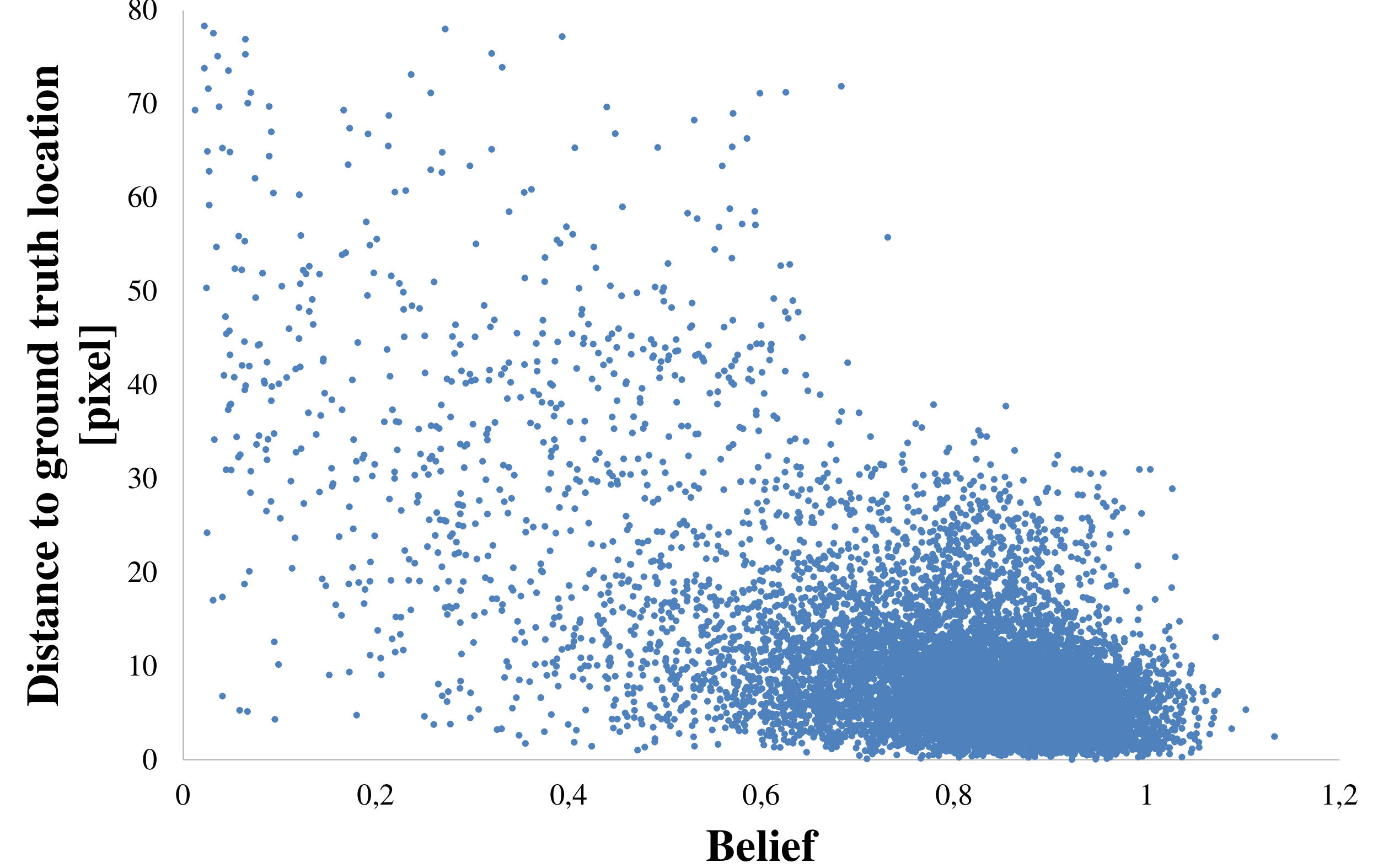}} 
\caption{a) detection accuracy in dependence of the viewing direction. b) correlation between belief and prediction error.}
\label{fig:detectionAccuracy}
\end{figure} 

\begin{figure}[tb] 
\centering
{\includegraphics[height=0.3\textwidth]{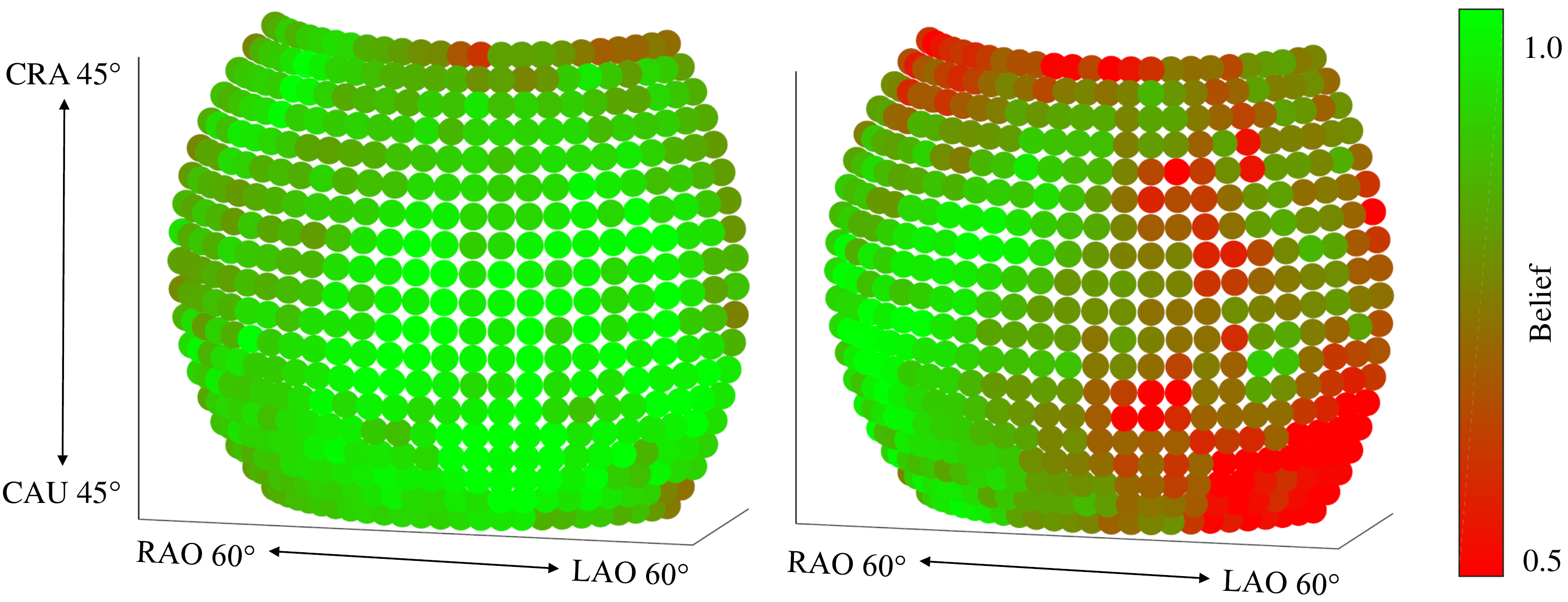}} 
\caption{Belief distribution of two single landmarks. Each landmark has it own detection belief distribution over the sphere.\label{fig:landmarkBelief}} 
\end{figure} 

\paragraph{Experiment} For evaluation, we uniformly sampled projections from our testing volume on a spherical segment covering the same angular range used in training. The angular increment between samples was \ang{5}, soure-to-isocenter distance was $750$\,mm, and source-to-detector distance was $1200$\,mm.

\paragraph{Confidence Development} In Fig.~\ref{fig:beliefMaps} the refinement of the belief maps is shown for two examples. After the first stage, several areas in the X-ray image have a high response due to the locality of feature detection. With increasing stages, the belief in the correct landmark location is increased. 

\paragraph{Qualitative Results} In Fig.~\ref{fig:wallOfIndependence}, example X-rays covering the whole angular range are shown. 
Visually, one notices very good overall agreement between the predicted and true landmark locations.

\paragraph{Belief Map Response and Prediction Error} The maximum belief is an indicator for the quality of a detected landmark. Fig.~\ref{fig:detectionAccuracy} shows the correlation between belief map response and prediction error. As motivated previously, we define a landmark as \textit{detected} if the maximum belief is $\geq 0.4$. Then, the mean prediction error with respect to ground truth is $9.1\pm7.4$ pixels ($5.6\pm4.5$\,mm).

\paragraph{View Invariance} The view invariance of landmark detection is illustrated in the spherical plot in Fig.~\ref{fig:detectionAccuracy}. We define accuracy as the ratio of landmarks with an error $<15$ pixels to all detected landmarks in that view.
The plot indicates that detection is slightly superior in AP compared to lateral views. To provide intuition on this observation, we visualize the maximum belief of two representative landmarks as a function of viewing direction in Fig~\ref{fig:landmarkBelief}. While the first landmark is robust to changes in viewing direction, the second landmark is more reliably detected in CAUD/RAO views.

\subsection{Real X-rays}

\begin{figure}[tb] 
\centering
{\includegraphics[width=1.0\textwidth]{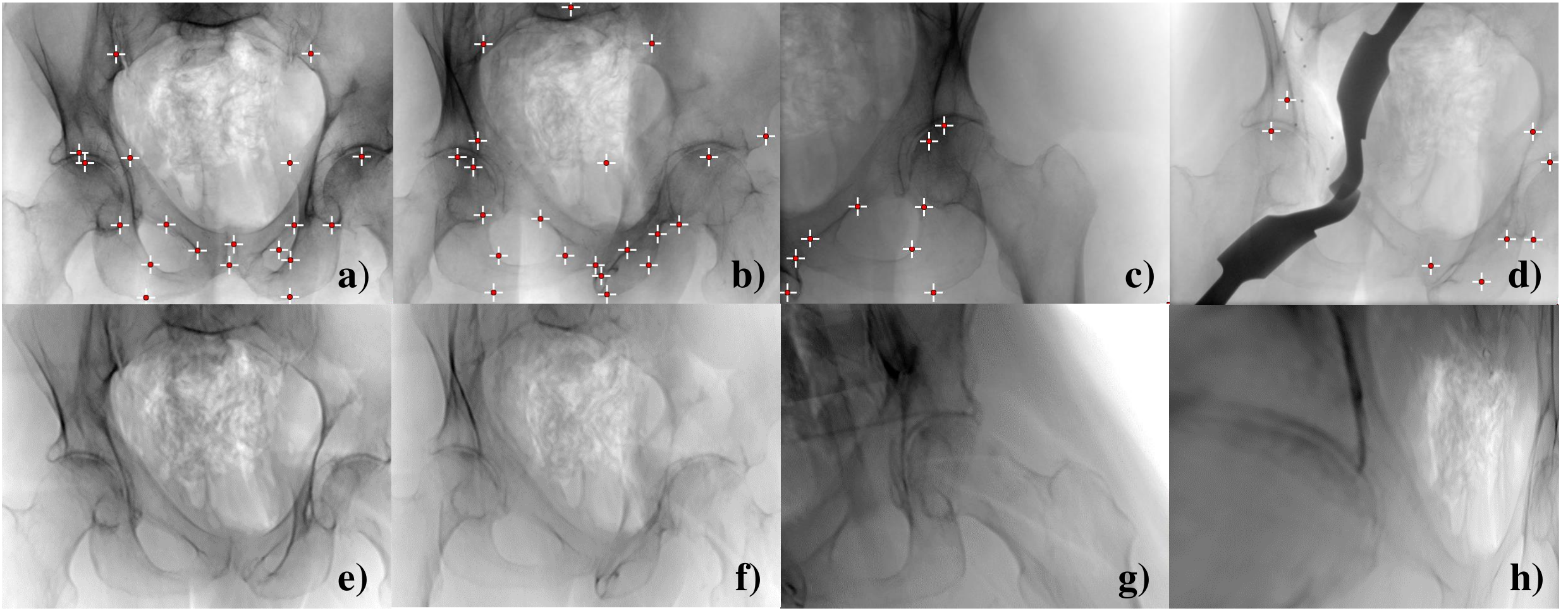}} 
\caption{Top: landmark predictions on clinical X-rays, indicated with a red label. Bottom: generated projection images after X-ray pose was retrieved using the detections.} 
\label{fig:realXray}
\end{figure} 

\paragraph{Landmark detection} Real X-ray images of a cadaver study were used to test generality of our model \emph{without} re-training. Sample images of our landmark detection are shown in Fig.~\ref{fig:realXray}, top row. Visually, the achieved predictions are in very good agreement with the expected outcome. Even in presence of truncation, landmarks on the visible anatomy are still predicted accurately, see Fig.~\ref{fig:realXray} c). A failure case of the network is shown in Fig.~\ref{fig:realXray} d), where a surgical tool in the field of view impedes landmark prediction.

\paragraph{Applications in 2D/3D Registration} As candidate clinical application, we study initialization of 2D/3D registration between pre-operative CT and intra-operative X-ray of cadaver data based on landmarks. Anatomical landmarks are manually labeled in 3D CT and automatically extracted from 2D X-ray images using the proposed method. Since correspondences between 2D detections and 3D references are known, the X-ray pose yielding the current view can be estimated in closed form~\cite{Hartley2004}. 
To increase robustness of the estimation and because belief of a landmark may depend on viewing direction, only landmarks with a belief above $0.7$ (but at least 6) are used. The estimated projection matrix is verified via forward projection of the volume in that geometry (Fig~\ref{fig:realXray}, bottom row). While initialization performs well in standard cases where most landmarks are visible and detected, performance deteriorates slightly in presence of truncation due to the lower amount of reliable landmarks, and exhibits poor performance if landmark detection is challenged by previously unseen scenarios, such as tools in the image. 

\section{Discussion and Conclusions}

We presented an approach to automatically detect anatomical landmarks in X-rays invariant of their viewing direction to benefit orthopedic surgeries by providing implicit 3D information.
Our results are very promising but some limitations remain. (1) As shown in Fig.~\ref{fig:realXray} d), the performance of our method is susceptible to scenarios not included in training, such as surgical tools in the image.(2) Lateral views of the pelvis exhibit slightly worse prediction performance compared to AP-like views. We attribute this behavior to more drastic overlap of the anatomy and lower amount of training samples seen by the network. We are confident that this effect can be compensated by increasing the angular range during training while limiting validation to the current range. Since some landmarks are equally well predicted over the complete angular range, the concept of maximum belief is powerful in selecting reliable landmarks for further processing. (3) Downsampling of ground truth belief map limits the accuracy of the detection despite efforts to increase accuracy, e.\,g. sub-pixel maximum detection.
Detecting anatomical landmarks proved essential in automatic image parsing in diagnostic imaging, but may receive considerable attention in image-guided interventions as new approaches, such as this one, strive for clinically acceptable performance. In addition to 2D/3D registration, we anticipate applications for the proposed approach in clinical tasks that inherently involve X-ray images from multiple orientations, in particular K-wire placement.



\end{document}